\documentclass[journal]{IEEEtran}
\usepackage{cite}

\ifCLASSINFOpdf
   \usepackage[pdftex]{graphicx}
\else
\fi
\usepackage{amsmath}
\usepackage{algorithm}
\usepackage{algorithmic}
 \usepackage[caption=false,font=footnotesize]{subfig}
\usepackage{multirow}
\newcommand{\tabincell}[2]{\begin{tabular}{@{}#1@{}}#2\end{tabular}}

\begin{document}

\title{Binarizing Weights Wisely for Edge Intelligence: \\Guide for Partial Binarization 
of Deconvolution-Based Generators}
%
%
%

\author{Jinglan~Liu, 
        Jiaxin~Zhang, 
        Yukun~Ding,
        Xiaowei~Xu,
        Meng~Jiang,
        and~Yiyu~Shi,~\IEEEmembership{Senior~Member,~IEEE}
\thanks{J. Liu, Y. Ding, X. Xu, M. Jiang, and 
Y. Shi are with the Department
of Computer Science and Engineering, University of Notre Dame, Notre Dame,
IN, 46556 USA e-mail: \{jliu16, yding5, xxu8, 
mjiang2, yshi4\}@nd.edu.}
\thanks{J. Zhang was with University of Science and Technology of China.}
}

%
%

\markboth{Journal of \LaTeX\ Class Files,~Vol.~14, No.~8, August~2015}%
{Shell \MakeLowercase{\textit{et al.}}: Bare Demo of IEEEtran.cls for IEEE Journals}
%



\maketitle

\begin{abstract}
This work explores the weight binarization of the deconvolution-based generator in a Generative Adversarial Network (GAN) for memory saving and speedup of image construction on the edge. 
Our study suggests that different from convolutional neural networks (including the discriminator) where all layers can be binarized, only some of the layers in the generator can be binarized without significant performance loss. Supported by theoretical analysis and verified by experiments, a direct metric based on the dimension of deconvolution operations is established, which can be used to quickly decide which layers in a generator can be binarized. Our results also indicate that both the generator and the discriminator should be binarized simultaneously for balanced competition and better performance during training. Experimental results on CelebA dataset with DCGAN and original loss functions suggest that directly applying state-of-the-art binarization techniques to all the layers of the generator will lead to 2.83$\times$ performance loss measured by sliced Wasserstein distance compared with the original generator, while applying them to selected layers only can yield up to 25.81$\times$ saving in memory consumption, and 1.96$\times$ and 1.32$\times$ speedup in inference and training respectively with little performance loss. Similar conclusions can also be drawn on other loss functions for different GANs. 
\end{abstract}


%
\IEEEpeerreviewmaketitle

\section{Introduction}
\IEEEPARstart{G}{enerative} 
adversarial networks (GANs), 
which are spin-offs from conventional 
convolutional neural networks (CNNs),
have attracted much attention in the 
fields of reinforcement learning, unsupervised 
learning and also semi-supervised learning 
\cite{finn2016unsupervised, pfau2016connecting,salimans2016improved}. 
A GAN is composed of two parts: 
a discriminator and a generator. 
Usually, discriminators are implemented 
by convolutional neural networks, while
generators are implemented by 
deconvolutional neural networks. 
More details about GANs will be 
presented in Section \ref{sec:gan}. 

\begin{figure}[htb]
    \centering
    \includegraphics[width=2.5in]{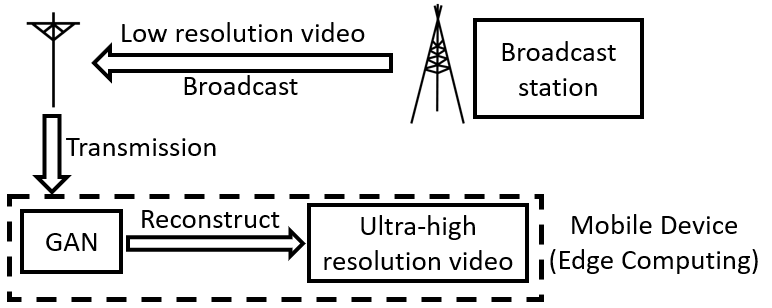}
    \caption{Low resolution broadcast based on GAN}
    \label{fig:broadcast}
\end{figure}

Some promising applications based on 
GANs include images reconstruction with 
super-resolution, art creation and 
image-to-image translation 
\cite{goodfellow2016nips}, 
many of which can run on mobile 
devices (edge computing). For example, 
one potential application of 
GANs allow videos to be broadcast 
in low resolution and then 
reconstructed to ultra-high 
resolution by end users 
\cite{ledig2016photo} as 
shown in Fig. \ref{fig:broadcast}. 

However, the resources required 
by GANs to perform computations 
in real-time may not be easily 
accommodated by mobile devices. 
For example, constructing an 
image of 64$\times$64 resolution 
with deep convolutional 
generative adversarial network 
(DCGAN) \cite{radford2015unsupervised} 
requires 86.6 MB of memory, 
most of which is used for 
the generator. 
The memory goes up to 620.8 MB for 
1024$\times$1024 resolution 
\cite{karras2017progressive}, 
and up to about 800 MB for 
the popular 4K video with 
resolution of 3840$\times$2160. 
On the other hand, one of the 
state-of-the-art mobile 
processors, A12 Bionic in the 
newest iPhone XS Max 
\cite{apple}, 
provides only 4 GB RAM, 
most of which must be 
occupied by the operating 
system and its peripheries. 
As a result, developers must 
restrict neural network models 
to just a few megabytes to 
avoid crash 
\cite{chen2015compressing}. 
The memory budget gets 
even tighter when it 
comes to mobile devices of 
smaller form factor 
such as Apple Watch series 3, 
which only has 768 MB RAM. 

The same problem has been well 
known for conventional CNNs, 
and various 
solutions have been proposed 
via redesigning the algorithms 
and/or computation structures 
\cite{han2015deep,zhang2016cambricon,guo2018angel,wang2017dlau}. 
Among them, quantization until 
binary is one of the most popular 
techniques as it fits hardware 
implementation well with high efficiency 
\cite{chen2015compressing,hubara2016quantized,al2018xnorbin}. However, quantization can reduce the expressive power of the neural networks significantly and the discrete parameters make the optimization much more difficult. Naive quantization usually leads to total failure especially for binarization. Significant effort has been devoted to develop better quantization and binarization methods as well as the hardware accelerator~\cite{li2016ternary,hubara2016quantized,zhang2018lq,conti2018xnor,ling2018taijinet}. Its success on CNNs has been 
demonstrated by multiple works, where memory consumption is deeply compressed although sometimes 
the performance cannot be preserved \cite{rastegari2016xnor,han2015learning,jouppi2017datacenter,zhou2016dorefa}. 

Compression techniques 
can be readily 
applied to discriminator networks 
in GANs, which are nothing 
different from conventional CNNs. 
It may be alluring to also apply 
the quantization techniques to binarize generators, 
especially the deconvolution-based 
\cite{zeiler2010deconvolutional} 
ones as the computation 
process looks similar. However, 
instead of distilling local 
information from a global map 
as in convolution operations, 
deconvolution attempts to construct 
the global map by local 
information. This difference 
can lead to significantly 
different binarization results, 
as will be discussed in Section 
\ref{sec:math}. 
Accordingly, a scheme tailored to 
deconvolution-based generators is 
warranted. 

In this paper, we show through 
theoretical analysis that under 
certain conditions, 
binarizing a deconvolution 
layer may cause significant 
performance loss, which also 
happens in compression of 
CNNs per empirical findings 
so far. Since there is no 
explanation for this phenomenon 
to the best of the authors' 
knowledge, an intuitive guess 
is that not all layers can 
be binarized together while 
preserving performance. 
Thus, some layers need to 
stay in the format of floating 
point for performance, while 
others can be binarized without 
affecting performance. 
To quickly decide whether 
a layer can be binarized, 
supported by theoretical 
analysis and verified by 
experiments, a simple yet 
effective metric 
based on the dimension of 
deconvolution operations is 
established. Based on this metric, 
we can make use of existing 
compression techniques to binarize 
the generator of GANs with little 
performance loss. 
We then propose 
the scheme of partial binarization 
of deconvolution-based 
generators (PBGen) under the 
guide of the metric. 

Furthermore, we find that only 
binarizing the generator and leaving 
discriminator network unchanged 
will introduce unbalanced 
competition and performance 
degradation. Thus, both networks 
should be binarized at the same 
time. Experimental results based 
on CelebA suggest that directly 
applying state-of-the-art 
binarization techniques to all 
the layers of the generator will 
lead to 2.83$\times$ performance 
loss measured by sliced Wasserstein 
distance compared with the original 
generator, while applying them to 
selected layers only can yield up 
to 25.81$\times$ saving in memory 
consumption, and 1.96$\times$ and 
1.32$\times$ speedup in inference 
and training respectively with little 
performance loss. 
The conclusions will stay the same even 
though different loss functions are utilized.

The remainder of the paper is organized as follows. 
Section \ref{sec:rela} discusses related works and 
background for compression techniques for CNN as well 
as GANs. Section \ref{sec:math} exhibits the theoretical 
analysis on the power of representation in 
deconvolution/convolution layers and the algorithm for 
model binarization based on it. Experiments 
for verification and performance are displayed in 
Section \ref{sec:exp}. This work is concluded in 
Section \ref{sec:conc}. 
\section{Related Works and Background}
\label{sec:rela}

\subsection{CNN Compression}
Compression techniques for CNNs 
mainly consist of 
pruning, quantization, re-structure and 
other approximations based on 
mathematical matrix manipulations 
\cite{han2015deep,jaderberg2014speeding,lin2013network}. 
The main idea of the 
pruning method in \cite{han2015learning} 
is to ``prune'' connections with 
smaller weights out so that both 
synapses and neurons are possible 
to be removed from the original 
structure. This can 
work well with traditional CNNs and 
reduce the number
of parameters of 
AlexNet by a factor of nine 
\cite{han2015learning}. 
Re-structure methods modify 
network structures for 
compression, such as 
changing functions or 
block order in layers 
\cite{lin2013network,rastegari2016xnor}. 

In this work, we focus 
on the quantization technique. 
Quantization aims to use fewer bits 
to present values of weights or even inputs. 
It has been used to accelerate 
CNNs in various works at different levels 
\cite{judd2016stripes,miyashita2016convolutional,xu2018quantization} 
including ternary quantization 
\cite{li2016ternary,zhu2016trained} and 
iterative quantization \cite{zhou2017incremental}, 
with small loss. 
In \cite{han2015deep}, the authors 
proposed to determine weight 
sharing after a network 
is fully trained, 
so that the shared 
weights approximate 
the original network. 
From a fully trained model, 
weights are clustered and 
replaced by the centroids of 
clusters. During 
retraining, the summation of the 
gradients in same groups are used for 
the fine-tuning of the centroids. Through such 
quantization, it is reported 
to be able to compress AlexNet 
up by around 8\% before 
significant accuracy loss occurs. 
If the compression rate goes 
beyond that, the accuracy will deteriorate rapidly. 

A number of recent works 
\cite{hubara2016quantized,zhou2016dorefa,rastegari2016xnor,cai2017deep,courbariaux2015binaryconnect,courbariaux2016binarized,ding2018universal} 
pushed it further 
by using binarization 
to compress CNNs, where 
only a single bit is used 
to represent values. 
Training networks with 
weights and activations 
constrained to $\pm$1 was 
firstly proposed in 
\cite{courbariaux2015binaryconnect}. 
Through transforming 32-bit 
floating point weight 
values to binary representation, 
CNNs with binary weights and activations 
are about 32$\times$ smaller. 
In addition, when weight values are 
binary, convolutions can be 
estimated by only addition 
and subtraction without 
multiplication, which can achieve 
around 2.0$\times$ speedup. 
However, the method introduces 
significant performance loss. 
To alleviate the problem, 
\cite{rastegari2016xnor} proposed 
Binary-Weight-Network, where 
all weight values are binarized 
with an additional continuous 
scaling factor for each output 
channel. We will base our 
discussion on this weight 
binarization afterwards, which is 
one of the state-of-the-art 
binarization methods.


Most recently, hybrid quantization has 
attracted more and more attention, because 
it enables better trade-off between 
compression and performance
~\cite{wang2018design,zhu2018adaptive,yin2018high}. 
As for partial binarization, a sub-area of 
hybrid quantization, on which we are focused, 
both training methods ~\cite{zhuang2017flexible} 
and the corresponding hardware 
accelerators~\cite{ling2018taijinet,xu2018efficient} are also 
investigated extensively. The actual performance 
after compression heavily depends on the 
configuration of the partial binarization, i.e. 
which layers are binarized while others are not. 
Given the fact that the search space is too 
large to do exhaustive search, finding the 
optimal or near-optimal configuration 
becomes a foremost challenge. 
\cite{chakraborty2019efficient} shows that 
adding full precision residual connections 
helps to reduce the loss of classification 
accuracy while getting excellent memory 
compression. One potential drawback of this 
method is that it introduces additional 
memory overhead. \cite{zhuang2017flexible} presents 
flexible network binarization with 
layer-wise priority, which is defined by 
the inverse of layer depth empirically. 
\cite{prabhu2018hybrid} proposes to use 
the binarization error of each layer as 
the indication of its importance to the 
final performance. They empirically show 
that partial binarization leads to 
significant improvement 
over fully binarized models. 
Very recently, \cite{chakraborty2019pca} 
also utilizes Principal Component Analysis (PCA) to 
identify significant layers in CNNs and uses 
higher precision on important layers. 
However, this method depends on pre-trained 
models, and PCA contains a large amount of 
computation naturally.



All things considered, none of the existing works 
explored the compression of generators in GANs, 
where deconvolution replaces convolution as 
the major operation. Note that while 
there is a recent work that uses the term of 
``binary generative adversarial networks'' 
\cite{song2017binary}, it is not about 
the binarization of GANs. In that work, 
only the inputs of the generator 
are restricted to binary code to 
meet the specific application requirement. 
All parameters inside 
the networks and the training 
images are not quantized. 

\subsection{GAN}
\label{sec:gan}
GAN was developed by 
\cite{goodfellow2014generative}
as a framework to train a generative 
model by an adversarial process. 
In a GAN, a discriminative 
network (discriminator) 
learns to distinguish whether 
a given instance is real 
or fake, and a generative 
network (generator) learns 
to generate realistic instances 
to confuse the discriminator. 

Originally, the discriminator 
and the generator of a GAN 
are both multilayer perceptrons. 
Researchers have since proposed many 
variants of it. For example, 
DCGAN transformed multilayer perceptrons 
to deep convolutional networks 
for better performance. Specifically, 
the generator is composed by four 
deconvolutional layers. 
GANs with such a 
convolutional/deconvolutional 
structure have also been 
successfully used to 
synthesize plausible visual 
interpretations of given 
text \cite{reed2016generative} 
and to learn interpretable and 
disentangled representation 
from images in an unsupervised 
way \cite{chen2016infogan}. 
Wasserstein generative 
adversarial networks (WGAN) 
\cite{arjovsky2017wasserstein} 
and least squares generative 
adversarial networks 
(LSGAN) \cite{mao2016least} 
have been proposed with
different loss functions 
to achieve more stable 
performance, yet they both 
employed the deconvolution 
operations too. To verify the 
robustness of our analysis, both 
DCGAN and LSGAN are tested 
in our experiments.

\section{Analysis on Power 
of Representation}
\label{sec:math}
In this section, 
to decide whether a layer 
can be binarized, 
we analyze the power of 
a deconvolution layer to 
represent any given 
mapping between the input and 
the output, and how such power 
will affect the performance after 
binarization.  
We will show that the performance 
loss of a layer is related to the 
dimension of the deconvolution, 
and develop a metric called the 
degree of redundancy to indicate 
the loss. 
Finally, based on the analysis, 
several inferences are deduced at 
the end of this section, 
which should lead to effective 
and efficient binarization. 

In the discussion below, 
we ignore batch normalization 
as well as activation operations 
and focus on the deconvolution 
operation in a layer, as only 
the weights in that operation 
are binarized. 
The deconvolution process 
can be transformed to equivalent 
matrix multiplication. 
Let $\mathbf{D}^I$ ($\in \mathcal{R}^
{{c_i}\times{h_i}\times{w_i}}$, 
where ${c_i}$, ${h_i}$ and ${w_i}$ 
are number of channels, height 
and  width of the input respectively) 
be the input matrix, 
and $\mathbf{D}^O$ 
($\in \mathcal{R}^
{{c_o}\times
{h_o}\times{w_o}}$, where ${c_o}$, 
${h_o}$ and ${w_o}$ are the 
number of channels, 
height and width of the output 
respectively) 
be the output matrix. 
Denote $\mathbf{K}$ 
($\in \mathcal{R}^{{c_i}\times{c_o}
\times{h_k}\times{w_k}}$, 
where ${h_k}$ and ${w_k}$ are the 
height and width of a kernel in 
the weight matrix) 
as the weight matrix to 
be deconvoluted with 
$\mathbf{D}^I$. 
Padding is ignored in the 
discussion, since it will not 
effect the results. 

For the deconvolution operation, 
the local regions in the output 
can be stretched out into columns, 
by which we can cast 
$\mathbf{D}^O$ to 
$\mathbf{D}^{Od}\in \mathcal{R}^{{s_i}
\times{r_o}}$, where 
$s_i = h_i  w_i, r_o = c_o  h_k  w_k$. 
Similarly, $\mathbf{D}^{Id}\in 
\mathcal{R}^{{s_i}
\times{c_i}}$ can be restructured 
from $\mathbf{D}^{I}$, and 
$\mathbf{K}^d\in 
\mathcal{R}^{c_i\times r_o}$ 
can be restructured from 
$\mathbf{K}$, where 
$s_i = h_i  w_i, r_o = c_o  h_k  w_k$. 
Please refer to \cite{course:cs231n} 
for details about the transform. 
Then, the deconvolution 
operation can be compactly written 
as 

\begin{equation}\label{equ:demul}
 \mathbf{D}^{Od} = \mathbf{D}^{Id} * \mathbf{K}^d, 
\end{equation}

\noindent where 
$*$ denotes matrix multiplication. 
$\mathbf{D}^{Id}$ and 
$\mathbf{D}^{Od}$ are the 
matrices containing pixels 
for an image or an intermediate 
feature map. During the 
training process, we adjust 
the values of $\mathbf{K}^d$ 
to construct a desired 
mapping between $\mathbf{D}^{Id}$ 
and $\mathbf{D}^{Od}$. 

We use $(\cdot)_j$ to denote the 
$j$-th column of a matrix. 
Then (\ref{equ:demul}) can 
be decomposed column-wise as 

\begin{equation}\label{equ:dechannel}
\mathbf{D}^{Od}_j = \mathbf{D}^{Id} * \mathbf{K}^d_j
,\quad 1\leq j\leq r_o, 
\end{equation}

\noindent 
where $\mathbf{K}^d_j\in \mathcal{R}^
{c_i}$ and $\mathbf{D}^{Od}_j\in 
\mathcal{R}^{s_i}$. 

Now we analyze a mapping between 
an arbitrary input $\mathbf{D}^{Id}$ and 
an arbitrary output $\mathbf{D}^{Od}_j$. 
From (\ref{equ:dechannel}), 
when the weights are continuously 
selected, all vectors that can be 
expressed by the right hand expression 
is a subspace $\Omega$ spanned by the 
columns of $\mathbf{D}^{Id}$, the 
dimension of which is $c_i$. 
Here we have assumed 
without loss of generality 
that $\mathbf{D}^{Id}$ has full column rank. 
When $c_i<s_i$, which is the dimension 
of the output space $\Phi$ where 
$\mathbf{D}^{Od}_j$ lies, 
$\Omega$ is of lower dimension 
than $\Phi$, 
and accordingly, $\mathbf{D}^{Od}_j$ 
can either be uniquely 
expressed as a linear combination 
of the columns in $\mathbf{D}^{Id}$ if it 
lies in $\Omega$ (i.e. a unique 
$\mathbf{K}^d_j$  exists), or cannot 
be expressed if it is not (i.e. no such 
$\mathbf{K}^d_j$ exists).  
When $c_i = s_i$, 
$\Omega$ and $\Phi$ are equivalent, 
and any $\mathbf{D}^{Od}$ can be 
uniquely expressed as a 
linear combination of the columns 
in $\mathbf{D}^{Id}$. 
When $c_i > s_i$, 
$\Omega$ and $\Phi$ are still 
equivalent, but any $\mathbf{D}^{Od}$ 
can be expressed as an infinite 
number of different
linear combinations of 
the columns in $\mathbf{D}^{Id}$. 
In fact, the coefficients 
$\mathbf{K}^d_j$ of these 
combinations lie in a 
$(c_i - s_i)$-dimensional 
sub-space $\Psi$. 

The binarization imposes a 
constraint on the possible values 
of the elements in $\mathbf{K}^d_j$. 
Only finite number of combinations 
are possible. 
If $c_i\leq s_i$, then at 
least one of these combinations 
has to be proportional to the unique 
$\mathbf{K}^d_j$ that yields the 
desired $\mathbf{D}^{Od}_j$ to preserve 
performance. If $c_i>s_i$, then 
one of these combinations needs 
to lie in the subspace $\Psi$ to 
preserve performance. Apparently, 
the larger the  dimension of $\Psi$ 
is, the more likely this will happen, 
and the less the performance loss is. 
A detailed math analysis 
is straightforward to illustrate 
this, and is omitted here in 
the interest of space. 
Accordingly, we define the 
dimension of 
$\Psi$, $c_i-s_i$, as the 
degree of redundancy in 
the rest of the paper. 
Note that when this metric 
is negative, it reflects 
that $\Omega$ is of lower 
dimension than $\Phi$ and 
thus this deconvolution 
layer is more vulnerable to 
binarization errors. 
In general, a higher 
degree of redundancy should 
give lower binarization error. 

\begin{figure}[htb]
\centering
\includegraphics[width=3.2in]{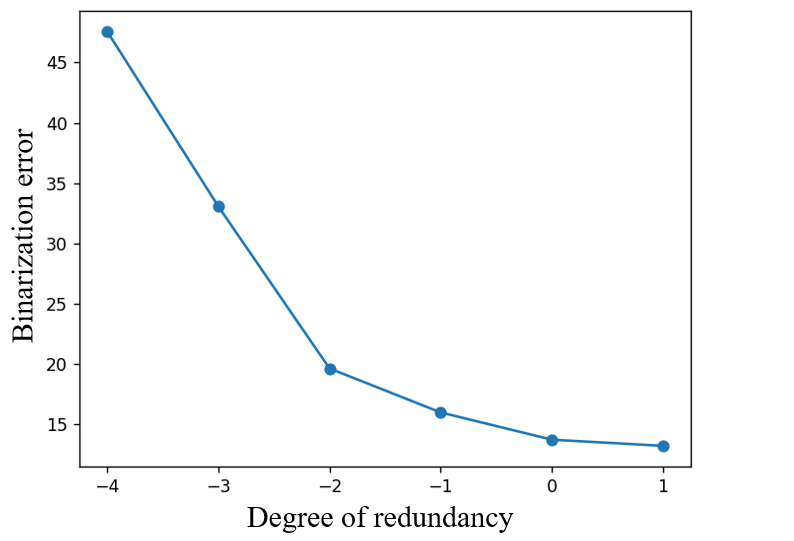}
\caption{Binarization error v.s. 
    degree of redundancy for a 
    deconvolution layer}
\label{fig:euc}
\end{figure}

We will use a small numerical 
example to partially validate 
the above discussion. We construct 
a deconvolution layer and vary 
its degree of redundancy by 
adjusting the $c_i$ in it, 
where $s_i = 20$.  
For each degree of redundancy 
we calculate the minimum average 
Euclidean distance between the 
original output and the 
output produced by binarized 
weights, 
which reflects the error 
introduced through the 
binarization process, 
referred to as binarization 
error throughout the paper.
The binarization error is obtained by 
enumerating all the possible 
combinations of those 
binary weights. The results are 
depicted in 
Fig. \ref{fig:euc}. 
From the figure we can 
see that the error decreases 
with the increase of degree of 
redundancy, which matches our conjecture. 

\begin{figure}[htb]
    \centering
    \includegraphics[width=3.2in]{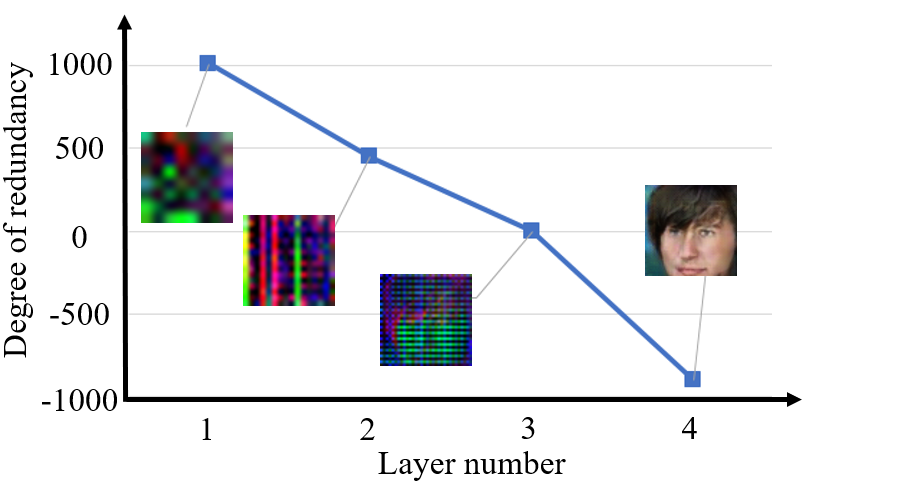}
    \caption{Degree of redundancy v.s. 
    layer number for DCGAN. The 
    intermediate feature maps at the 
    output of each layer as well as 
    the final output are also presented }
    \label{fig:lvsredun}
\end{figure}

For generators in most 
state-of-the-art GAN models 
\cite{radford2015unsupervised,mao2016least}, 
we find that the degree of redundancy 
reduces with the increase in depth, 
eventually dropping below zero. 
Such a decrease 
reflects the fact that more details 
are generated at the output of a 
layer as its depth grows, as can 
also be seen in Fig. \ref{fig:lvsredun}. 
These details are highly correlated, 
and reduce the subspace needed to 
cover then. 

Based on our analysis, several 
inferences can be deduced 
to guide the binarization: 
\begin{itemize}
\item With the degree of redundancy, 
    taking advantage of existing 
    binarization methods becomes 
    reasonable and feasible. Binarizing 
    layers with higher degree of 
    redundancy will lead to lower 
    performance loss after binarization, 
    while layers 
    with negative degree of 
    redundancy should be kept 
    un-binarized to avoid excessive 
    performance loss. 
\item According to the chain rule 
    of probability in directed graphs, 
    the output of every layer is only 
    dependent on its direct input. 
    Therefore, the binarizability 
    of each layer can be superposed. If a 
    layer can be binarized alone, it 
    can be binarized with other such 
    layers. 
\item When binarizing several deconvolution 
    layers together, the layer with the 
    least degree of redundancy may be the 
    bottleneck of the generator's 
    performance. 
\end{itemize}
As a result, only shallower layer(s) 
of a generator can be binarized 
together to preserve its performance, 
because of the degree of redundancy 
trend in it. This leads to PBGen. 
Besides, such analysis may also explain 
why binarization can be applied in 
almost all convolution layers: 
distilling local information from 
a global map leads to positive 
degree of redundancy. 

Under the guide of such inferences, 
the algorithm for wise binarization 
on Deconv/Conv networks can be implemented 
by calculating the degree of redundancy for 
every Deconv/Conv layer at first; then all 
these layers can be sorted by their degree of 
redundancy from high to low; in this order, 
every layer will be binarized singly to observe 
the difference in performance from the 
original full-precision version and whether 
to continue or not will be decided by the 
trade-off between performance and efficiency; 
finally, selected layers will be binarized 
together to serve as the ultimate strategy 
for network binarization. This algorithm 
can be described as 
Algorithm \ref{alg}.
We attempt to preserve the original 
full-precision performance in our 
experiments. 

\begin{algorithm}
\caption{Wise binarization on 
Deconv/Conv networks}
\label{alg}
\begin{algorithmic}
\STATE $L\gets$ number of Deconv/Conv layers
\STATE $T\gets$ performance degradation threshold
\FOR{$l=1:L$}
    \STATE{Compute the degree of redundancy of each layer $R_l$}
\ENDFOR
\STATE{$\mathbf{R} \gets\ $sorted $R_1\cdots,R_L$ from high to low}
\STATE{$i\gets 1$}
\WHILE{$i\leq L$}
    \STATE{Binarize the $i$-th layer}
    \IF{performance degradation exceeds $T$}
        \STATE{\textbf{break}}
    \ENDIF
    \STATE{$i\gets i+1$}
\ENDWHILE
\RETURN $i$
\STATE{The network binarization strategy is 
to binarize the first $i-1$ layers according to $\mathbf{R}$ together}
\end{algorithmic}
\end{algorithm}

In addition, 
following the same derivation process for 
deconvolution in our manuscript, the degree of
redundancy of 
a convolution layer can be defined as 
$w_k\times h_k\times c_i - c_o$, instead 
of $c_i - s_i$ for deconvolution. 
Usually in a convolution layer, 
$c_o = 2\times c_i$, 
$w_k = h_k$, and $w_k = 3$ or $5$. 
As a result, the degree of redundancy 
is usually positive, 
and convolution is more readily binarizable 
compared to deconvolution as well. This also 
explains why using $1\times 1$ kernels 
will help compress networks while not hurting 
the networks' performance \cite{iandola2016squeezenet}.
\begin{figure}
    \centering
    \includegraphics[width=3.2in]{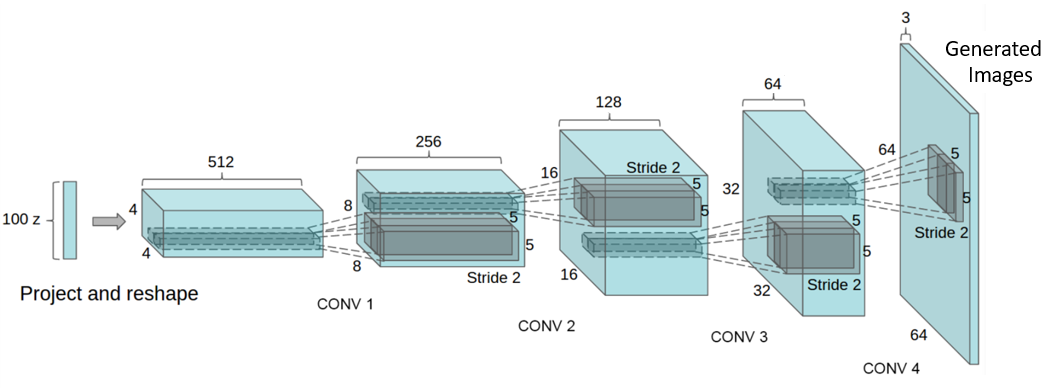}
    \caption{Structure of the generators in DCGAN
    with dimension of each layer labeled. 
    Deconvolutional layers are denoted 
    as ``CONV'' (figure credit: \cite{radford2015unsupervised})}
    \label{fig:struct_g4}
\end{figure}

\section{Experiments}
\label{sec:exp}

\subsection{DCGAN and Different 
Settings}
DCGAN will serve as a vehicle 
to verify the inferences 
deducted from the theoretical 
analysis in Section 
\ref{sec:math}. Except for the 
original adversarial loss function 
for DCGAN, least square loss function 
proposed in LSGAN is also tested in 
our experiments. The least square loss 
is one of the most popular loss functions 
for GANs, because it has been 
proved to be efficient for 
different GANs training.

We will explore 
how to best binarize it with 
preserved performance. 
Specifically, we use the 
TensorFlow \cite{abadi2016tensorflow} 
implementation of DCGAN on GitHub 
\cite{github:DCGAN}. The structure 
of its generator is illustrated in 
Fig. \ref{fig:struct_g4}. 
The computed degree of redundancy 
in each layer in the generator is shown 
in Fig. \ref{fig:lvsredun} and 
qualitatively summarized 
in TABLE \ref{tlb:redun}. 
The degree of redundancy in the last 
layer drops to -960. 
According to the inferences 
before, we can expect that since the 
degree of redundancy decreases as 
the depth increases, binarizing shallower 
layers and keeping the deeper layers in 
the format of floating point will 
help preserve the performance. 
For the readers' information, the degrees 
of redundancy of the four convolution 
layers in the discriminator are 11, 1472, 
2944, and 5888 respectively. 
\begin{table}[htb]
  \begin{center}
  \caption{Degree of redundancy in each 
  deconvolution layer of the generator in DCGAN}
  \label{tlb:redun}
  \begin{tabular}{c|c|c}
  \hline
  Layer number & Label in 
  Fig. \ref{fig:struct_g4} 
  & Degree of redundancy \\
  \hline
  1 & CONV 1 & 496 \\
  \hline
  2 & CONV 2 & 192 \\
  \hline
  3 & CONV 3 & -128 \\
  \hline
  4 & CONV 4 & -960 \\
  \hline
  \end{tabular}
  \end{center}
\end{table}

The binarization method used in 
Binary-Weight-Networks (BWN) 
proposed in \cite{rastegari2016xnor} 
is adopted to binarize layers 
no matter in a generator network 
or in a discriminator network. 
In BWN, all the weight values 
are approximated with binary 
values. Through keeping 
floating-point gradients while 
training, BWN is able to trained 
from scratch without pre-train.

There are four deconvolution 
layers in total in the generator, 
and each layer can be either 
binarized or not. 
For verification, 
we have conducted 
experiments on all $2^4=16$ 
different settings, but only the 
eight representative ones are 
discussed for clarity and space, 
and others will lead us to the 
same conclusion. Those 
eight different representative 
settings are summarized in 
TABLE \ref{tlb:setting} for clearness. 
In this table, 
the ``Setting'' column labels 
each setting. 
``Layer(s) binarized'' 
indicates which layer(s) are 
binarized in the generator. 
The ``Discriminator binarized'' 
column tells whether 
the discriminator is 
binarized or not. ``Y'' means yes, 
while ``N'' means no. This 
column is introduced to 
verify an observation in 
our experiments to be 
discussed later. Although 
settings in experiments 
include unbinarized discriminator 
and binarized discriminator, 
whether the discriminator is 
binarized or not will not affect 
the generated images significantly. 
That is, if the generator cannot 
generate recognizable faces with 
the unbinarized discriminator, 
it still cannot generate any 
recognizable faces with a binarized 
discriminator; and vice versa.
\begin{table}
  \begin{center}
  \caption{Settings of different partial 
  binarization of generator in 
  DCGAN}
  \label{tlb:setting}
  \begin{tabular}{c|c|c}
  \hline
  Setting & \tabincell{c}{Layer(s) binarized} & \tabincell{c}{Discriminator\\binarized}\\
  \hline
  A &  None & N \\
  \hline
  B & 1 & N \\
  \hline
  C & 2 & N \\
  \hline
  D & 3 & N \\
  \hline
  E & 4 & N \\
  \hline
  F & 1,2,3 & N \\
  \hline
  G & 1,2,3,4 & N \\
  \hline 
  H & 1,2,3 & Y \\
  \hline
  \end{tabular}
  \end{center}
\end{table}

Setting G will serve as 
the baseline model for performance 
after binarization, 
because it adopts the compression 
techniques based on CNNs directly 
without considering the degree 
of redundancy. 
On the other 
hand, Setting A serves as the 
baseline model when considering 
the memory saving, speedup as 
well as performance difference 
before and after binarization, 
because it 
represents the original DCGAN 
in floating point representation. 
It is considered as one common 
GAN structure providing good 
performance. 

\subsection{Dataset and Metrics}
CelebA \cite{liu2015faceattributes} 
is used as the dataset for 
our experiments, because it 
is a popular and verified dataset 
for different GAN structures. DCGAN, 
WGAN, LSGAN and many other GAN 
structures are tested on it 
\cite{github:gans-comp}. As 
every image in CelebA contains 
only one face, it is much easier to 
tell the quality of the generated 
images. 

Traditionally the quality of 
the generated images is 
identified by observation. 
However, qualitatively evaluation 
is always a hard problem. 
According to the in-depth analysis of 
commonly used criteria in 
\cite{theis2016anote}, 
good performance in a single or 
extrapolated metric from 
average log-likelihood, 
Parzen window estimates, 
and visual fidelity of samples does 
not directly translate to 
good performance of a GAN. 
On the other hand, the log-likelihood 
score proposed in \cite{wu2016quantitative} 
only estimates a lower bound 
instead of the actual performance. 

\begin{figure*}[htb]
\subfloat[Setting A]{
\begin{minipage}[b]{0.25\linewidth}
\label{fig:orig}
\includegraphics[width=1.5in]{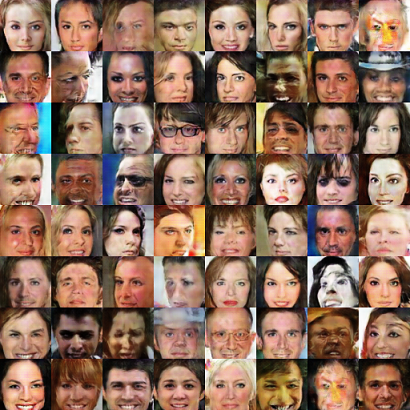}
\end{minipage}
}
\subfloat[Setting B]{
\begin{minipage}[b]{0.25\linewidth}
\label{fig:g1}
\includegraphics[width=1.5in]{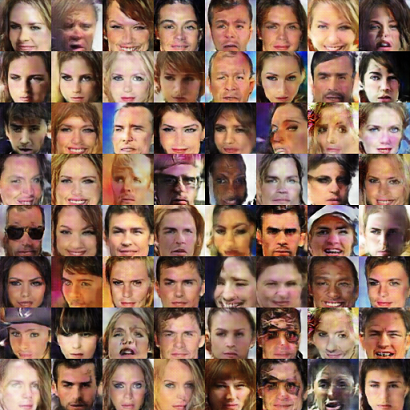}
\end{minipage}
}
\subfloat[Setting C]{
\begin{minipage}[b]{0.25\linewidth}
\label{fig:g2}
\includegraphics[width=1.5in]{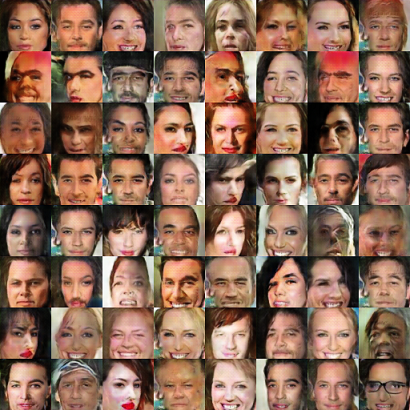}
\end{minipage}
}
\subfloat[Setting D]{
\begin{minipage}[b]{0.25\linewidth}
\label{fig:g3}
\includegraphics[width=1.5in]{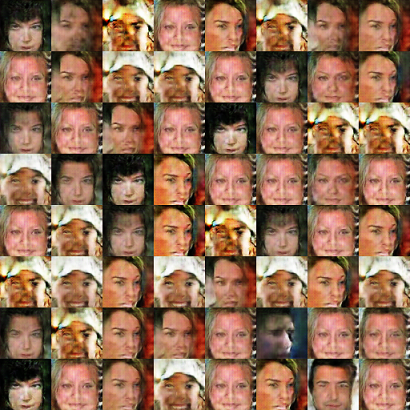}
\end{minipage}
}
\\
\subfloat[Setting E]{
\begin{minipage}[b]{0.25\linewidth}
\label{fig:g4}
\includegraphics[width=1.5in]{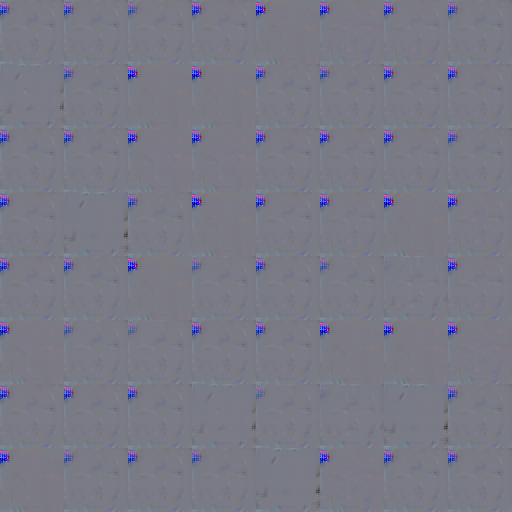}
\end{minipage}
}
\subfloat[Setting F]{
\begin{minipage}[b]{0.25\linewidth}
\label{fig:g1g2g3y}
\includegraphics[width=1.5in]{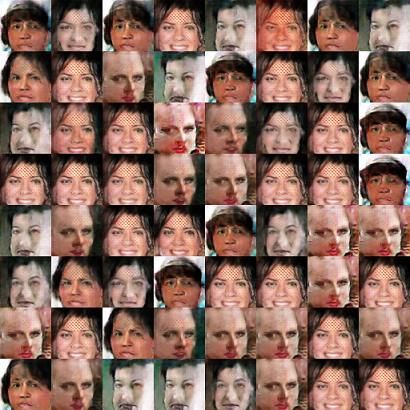}
\end{minipage}
}
\subfloat[Setting G]{
\begin{minipage}[b]{0.25\linewidth}
\label{fig:g1g2g3g4}
\includegraphics[width=1.5in]{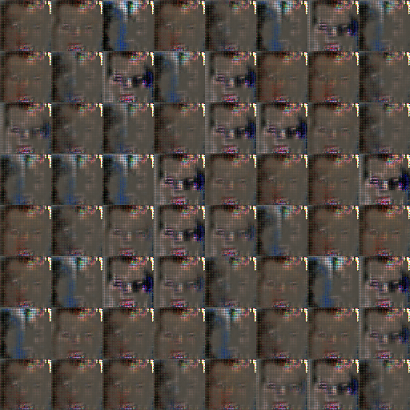}
\end{minipage}
}
\subfloat[Setting H]{
\begin{minipage}[b]{0.25\linewidth}
\label{fig:d4g3}
\includegraphics[width=1.5in]{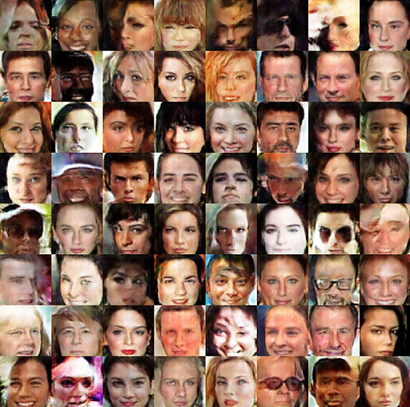}
\end{minipage}
}
\caption{Images generated under 
different settings}
\label{fig:deconv}
\end{figure*}

Very recently, 
\cite{karras2017progressive} 
proposed an efficient metric, 
which we will use in our 
experiments, and showed that 
it is superior to MS-SSIM 
\cite{odena2016conditional}, 
which is a commonly used 
metric. 
It calculates the sliced 
Wasserstein distance (SWD) 
between the training samples and the 
generated images under different 
resolutions. In particular, the SWD 
from the lower resolution 
patches indicates similarity in 
holistic image structures, while 
the finer-level patches encode 
information about pixel-level 
attributes. 
In this work, 
the max resolution is 64$\times$64. 
Thus, according to 
\cite{karras2017progressive}, 
we will use three different 
resolutions to evaluate the 
performance: 16$\times$16, 
32$\times$32, and 64$\times$64. 
For all different resolutions, 
small SWD indicates that the 
distributions of the patches 
are similar, which means that 
a generator with smaller SWD 
is expected to produce images 
more similar to 
the images from the training 
samples 
in both appearance and variation. 

\subsection{Experimental Results}
\label{sec:expr}
In this section, we will present 
experimental results that 
verify our inferences in Section 
\ref{sec:math}, along with some 
additional observations about 
the competition between the 
generator and the discriminator. 
The images generated by 
the original GAN (Setting A), 
in which all weights of each deconvolution 
layer are in the form of floating 
point, are displayed in Fig. 
\ref{fig:orig}. The images 
generated by the binarized DCGAN 
without considering the degree of 
redundancy are displayed in 
Fig. \ref{fig:g1g2g3g4}. 
These are our two baseline 
models.

\subsubsection{Qualitative Comparison of 
Single-Layer Binarization }
We start our experiments by comparing 
the images generated by binarizing a 
single layer in the generator 
of DCGAN. The results are shown in 
Figures \ref{fig:g1} - \ref{fig:g4}, 
which are generated by PBGen's under 
Setting B - Setting E respectively. In 
other words, those PBGen's utilize 
binary weights to the first, 
the second, the third, and the last 
deconvolution layer respectively. 
The degree of redundancy of each layer 
is shown in Fig. \ref{fig:lvsredun}. 
From the generated figures we can then see 
that Fig. \ref{fig:g1} generates 
the highest quality of images, 
similar to the original ones in 
Fig. \ref{fig:orig}. Images in 
Fig. \ref{fig:g2} are slightly 
inferior to those in Fig. \ref{fig:g1}, 
but better than those in Fig. \ref{fig:g3}. 
Fig. \ref{fig:g4} has no meaningful 
images at all. These observations 
are in accordance with our inferences 
in Section \ref{sec:math}: 
the performance loss when binarizing 
a layer is decided by its degree 
of redundancy, and a layer with 
negative degree of redundancy 
should not be binarized. 

\begin{figure}
\centering
\includegraphics[width=2.9in]{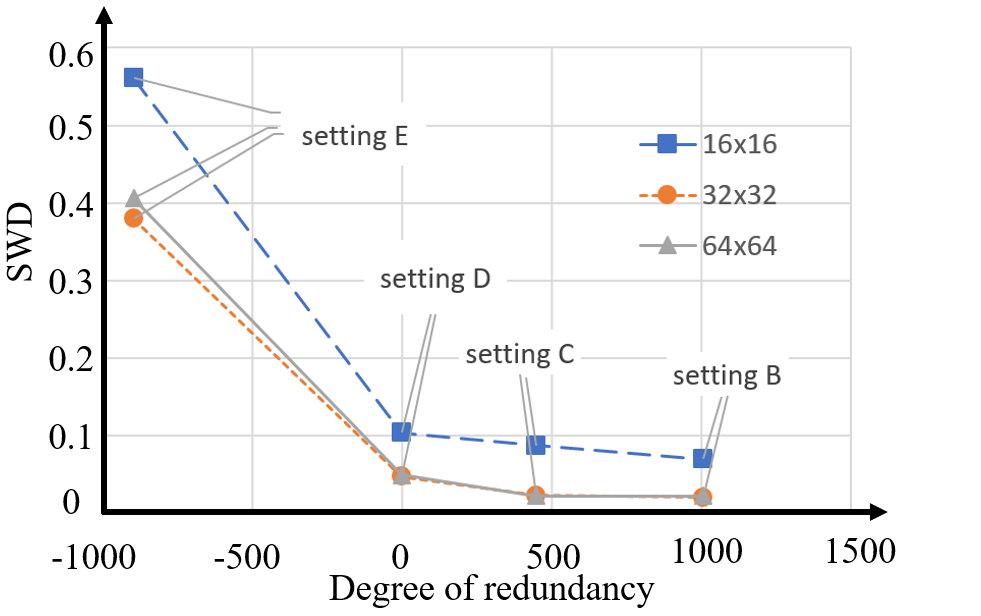}
\caption{SWD v.s degree of redundancy of 
         the binarized layer 
         in different settings}
\label{fig:dvsswd}
\end{figure}

\begin{figure}
\centering
\subfloat[64]{
\begin{minipage}[b]{0.16\linewidth}
\label{fig:chn64}
\includegraphics[width=0.5in]{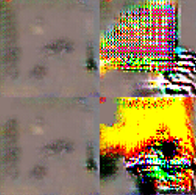}
\end{minipage}
}
\subfloat[128]{
\begin{minipage}[b]{0.16\linewidth}
\label{fig:chn128}
\includegraphics[width=0.5in]{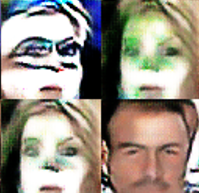}
\end{minipage}
}
\subfloat[256]{
\begin{minipage}[b]{0.16\linewidth}
\label{fig:chn256}
\includegraphics[width=0.5in]{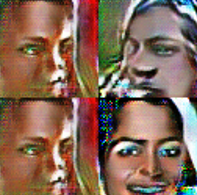}
\end{minipage}
}
\subfloat[512]{
\begin{minipage}[b]{0.16\linewidth}
\label{fig:chn512}
\includegraphics[width=0.5in]{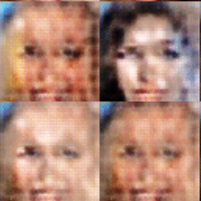}
\end{minipage}
}
\subfloat[1024]{
\begin{minipage}[b]{0.16\linewidth}
\label{fig:chn1024}
\includegraphics[width=0.5in]{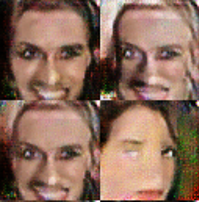}
\end{minipage}
}

\caption{Generated images of faces under Setting E with different number of input channels for the CONV4 layer. The number of the CONV4 layer for each experiments is (a) 64, (b) 128, (c) 256, (d) 512, and (e) 1024 respectively. With the increased number of 
input channels in the CONV4 layer, the degree of redundancy of this layer increases while other layers' degree of redundancy stays the same. This validates the degree of redundancy as an indication of the capability for a layer.}
\label{fig:diff_chn}
\end{figure}

To address the concern that low 
performance of Setting E is 
caused by the low degree of redundancy instead of 
the position of the layer (the last 
layer), more experiments 
are conducted under Setting E. As 
analysed in Section~\ref{sec:math}, 
degree of redundancy is defined by $c_i-s_i$, thus 
changing $c_i$ of one layer will only 
change the degree of redundancy of that 
layer, and will not have an effect on 
other layers' degree of redundancy. 
Thus, experiments with different 
number of input channels for the 
CONV4 layer, $c_4$, under Setting E. 
The generated images of faces 
in these experiments are 
shown in Fig.~\ref{fig:diff_chn}. 
The original $c_4$ is 64, and experiments 
are also conducted with 128, 256, 512, 
and 1024 respectively. 
According to TABLE~\ref{tlb:redun}, 
the degree of redundancy of 
CONV4 is zero when $c_4 = 1024$. 
As shown in Fig.~\ref{fig:diff_chn}, 
the generated images get clearer with 
more details along with the increased 
number of input channels and higher 
degree of redundancy.

\begin{figure}[h]
\centering
\subfloat[None]{
\begin{minipage}[b]{0.16\linewidth}
\label{fig:b0}
\includegraphics[width=0.5in]{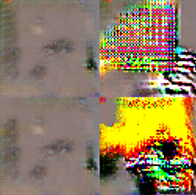}
\end{minipage}
}
\subfloat[CONV1]{
\begin{minipage}[b]{0.16\linewidth}
\label{fig:b1}
\includegraphics[width=0.5in]{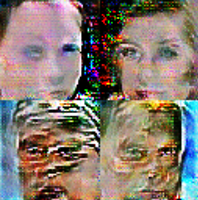}
\end{minipage}
}
\subfloat[CONV2]{
\begin{minipage}[b]{0.16\linewidth}
\label{fig:b2}
\includegraphics[width=0.5in]{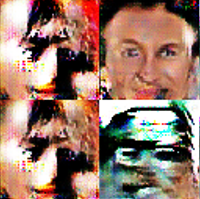}
\end{minipage}
}
\subfloat[CONV3]{
\begin{minipage}[b]{0.16\linewidth}
\label{fig:b3}
\includegraphics[width=0.5in]{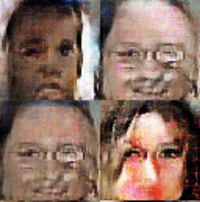}
\end{minipage}
}
\subfloat[CONV4]{
\begin{minipage}[b]{0.16\linewidth}
\label{fig:b4}
\includegraphics[width=0.5in]{images/b4.png}
\end{minipage}
}
\caption{Generated images of faces after binarizing the CONV4 layer. The layer with increased DOR for each experiment is (a) none, (b) CONV1, (c) CONV2, (d) CONV3, and (e) CONV4 respectively. The DOR is increased by 960 for each experiment except for (a). We can see that increasing the DOR of other layers cannot solve the bottleneck problem introduced by CONV4, but increasing the DOR of CONV4 can.}
\label{fig:diff_layer}
\end{figure}

In addition, we also conducted the experiments 
that vary the DOR of the first three layers when 
binarizing the fourth layer to validate that 
every layer binarization is relatively independent. 
In fact, none of these experiments could 
achieve the same performance improvement 
as that when increasing the DOR of the fourth 
layer by a same number. Generated images are shown 
in Fig. \ref{fig:diff_layer}. 

\subsubsection{Quantitative Comparison 
of Single-Layer Binarization} 
We further quantitatively compute 
the SWD values with 16$\times$16, 
32$\times$32 and 64$\times$64 
resolutions for Setting B, Setting C, 
Setting D, and Setting E. Their 
relationship with the degree of redundancy 
of the binarized layer 
is plotted in Fig. \ref{fig:dvsswd}. 
From the figure, two things are clear: 
first, regardless of resolution, 
a negative degree of redundancy 
(Setting E) results in a more 
than 5$\times$ increase in SWD 
compared with other settings 
with non-negative degree of redundancy 
(Setting B, Setting C, and Setting D). 
Second, for all the three resolutions, 
SWD decreases almost linearly with 
the increase of the degree of 
redundancy when it is non-negative. 
This confirms that our degree 
of redundancy can capture 
the impact of binarization 
not only on the holistic 
structure but also on the 
pixel-level fine details, 
and as such, is indeed a good 
indicator to quickly judge 
whether a layer can be binarized. 

We also report the SWD averaged 
over different resolutions 
(16$\times$16, 32$\times$32, 
64$\times$64) in TABLE 
\ref{tlb:swd}, where the result 
for the original GAN (Setting A) 
is also reported. 
From the table we can draw 
similar conclusions, that binarizing 
second layer (Setting C) increases 
the average SWD by 2.3\% compared 
with the original GAN (Setting A), 
while binarizing 
third and fourth layer 
(Setting D and Setting E) further 
increases it by 52.3\% and 913.6\%, 
respectively. 
\begin{figure}[h]
\centering
\includegraphics[width=3in]{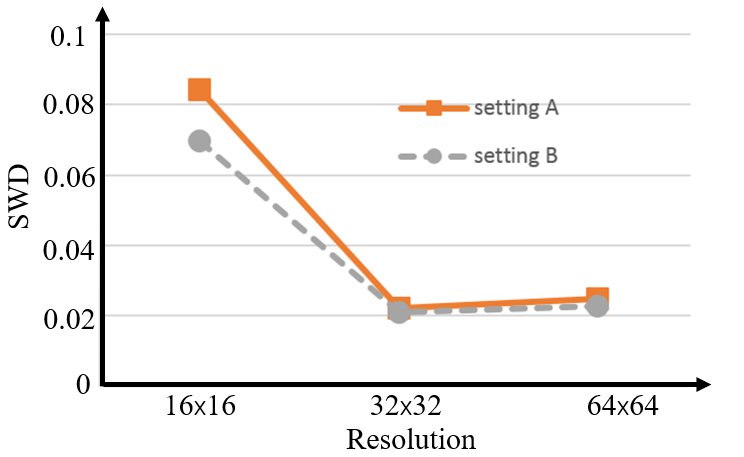}
\caption{SWD v.s. resolution for SWD 
         score calculation under 
         Setting A and Setting B}
\label{fig:swd_reso_ab}
\end{figure}
\begin{table}[bht]
  \begin{center}
  \caption{Average SWD under different settings}
  \label{tlb:swd}
  \begin{tabular}{c|c|c|c|c|c}
  \hline
  Setting & A & B & C & D & E \\
  \hline
  \ Average SWD ($\times 10^{-3}$) \ & \ 44 \ &
  \ 38 \ & \ 45 \ & \ 67 \ & \ 449 \ \\
  \hline
  \end{tabular}
  \end{center}
\end{table}

It is interesting to note that 
the average SWD achieved by 
binarizing the first layer (Setting B) 
is 13.6\% smaller than that from the 
original DCGAN (Setting A). To further 
check this, we plot the SWD v.s. 
resolution for these two settings 
in Fig. \ref{fig:swd_reso_ab}. 
From the figure we 
can see that the SWD from Setting B 
is always smaller than that from 
Setting A across all three resolutions. 
This shows that Setting B can achieve 
better similarity as well as 
detailed attributes. Such an improvement is 
probably due to the regularization 
effect, and similar effect has been 
observed in the compression of CNNs 
\cite{cai2017deep}. 

\begin{figure*}[htb]
\subfloat[Setting A]{
\begin{minipage}[b]{0.25\linewidth}
\label{fig:orig_lsgan}
\includegraphics[width=1.5in]{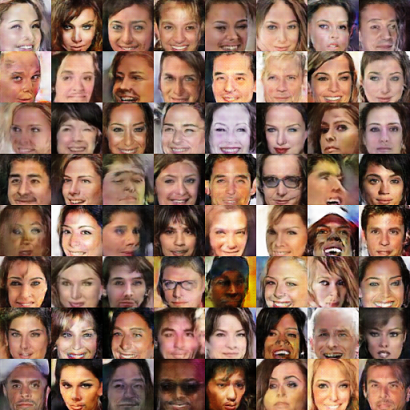}
\end{minipage}
}
\subfloat[Setting B]{
\begin{minipage}[b]{0.25\linewidth}
\label{fig:g1_lsgan}
\includegraphics[width=1.5in]{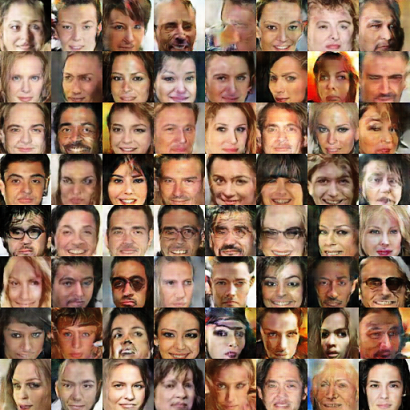}
\end{minipage}
}
\subfloat[Setting C]{
\begin{minipage}[b]{0.25\linewidth}
\label{fig:g2_lsgan}
\includegraphics[width=1.5in]{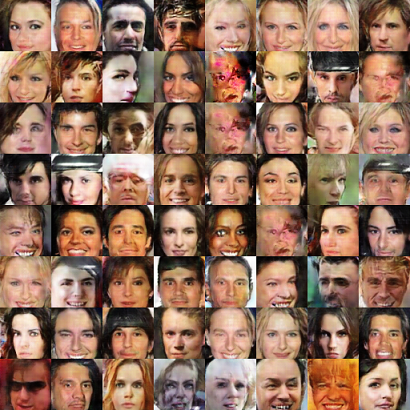}
\end{minipage}
}
\subfloat[Setting D]{
\begin{minipage}[b]{0.25\linewidth}
\label{fig:g3_lsgan}
\includegraphics[width=1.5in]{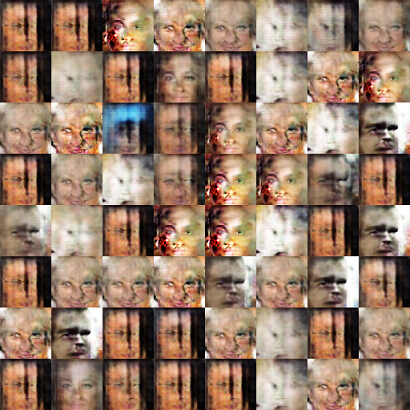}
\end{minipage}
}
\\
\subfloat[Setting E]{
\begin{minipage}[b]{0.25\linewidth}
\label{fig:g4_lsgan}
\includegraphics[width=1.5in]{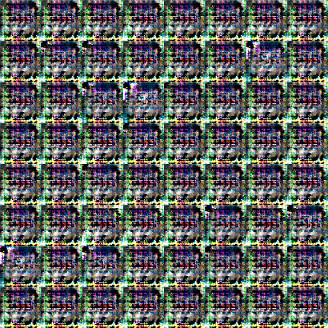}
\end{minipage}
}
\subfloat[Setting F]{
\begin{minipage}[b]{0.25\linewidth}
\label{fig:g1g2g3_lsgan}
\includegraphics[width=1.5in]{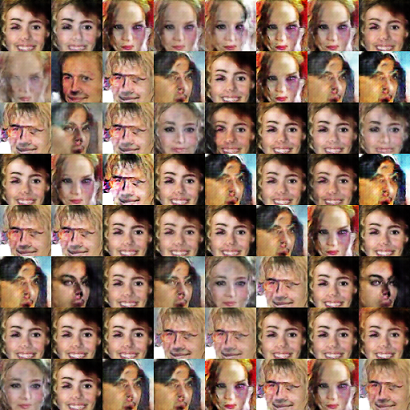}
\end{minipage}
}
\subfloat[Setting G]{
\begin{minipage}[b]{0.25\linewidth}
\label{fig:g1g2g3g4_lsgan}
\includegraphics[width=1.5in]{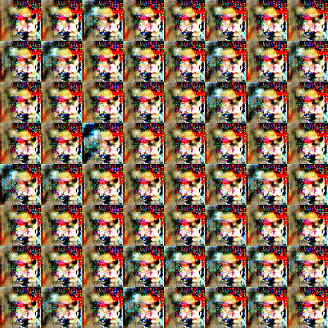}
\end{minipage}
}
\subfloat[Setting H]{
\begin{minipage}[b]{0.25\linewidth}
\label{fig:d4g3_lsgan}
\includegraphics[width=1.5in]{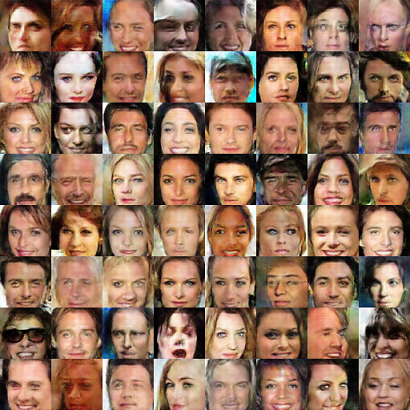}
\end{minipage}
}
\caption{Images generated under 
different settings using the least square loss}
\label{fig:deconv_lsgan}
\end{figure*}

\subsubsection{Validation of 
Superposition of Binarizability}
We now explore experiments to 
verify our inference that all 
layers that can be binarized 
alone can be binarized together. 
The images generated by Setting 
F in Fig. \ref{fig:g1g2g3y}, 
where the first three layers in 
the generator are binarized together, 
show no significant difference 
from those in Figures \ref{fig:orig}-
\ref{fig:g3}. Binarizing any 
two layers from the first three 
layers (not shown here) 
will lead to the same 
result. 
On the other hand, Setting 
G does not generate 
any meaningful output (Fig. 
\ref{fig:g1g2g3g4}), as the last 
layer, which cannot be binarized 
alone, is binarized together 
with the first three layers. 
Binarizing any of the first 
three layers as well as the 
last layer (not shown here) 
will produce 
meaningless results too. 
Setting G follows 
the state-of-the-art binarization 
for CNNs directly without 
considering the degree 
of redundancy. 
That is, with the assistance of 
the degree of redundancy, we can 
figure out that at most 
the first three 
deconvolution layers can 
be binarized with small loss 
on performance in the generator 
(Setting F). 
Nevertheless, directly adopting 
the existing binarization method 
will lead to excess degradation 
in performance and cannot provide 
any hint to improve (Setting G). 

\begin{figure}[htb]
    \centering
    \includegraphics[width=3in]{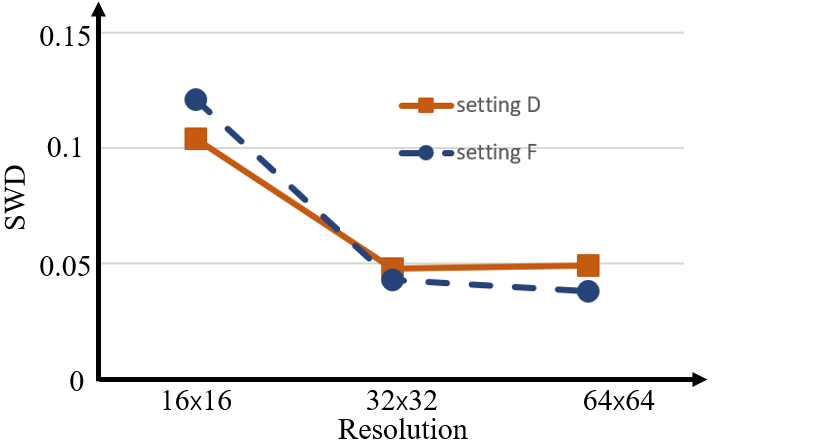}
    \caption{SWD v.s. resolutions 
    under Setting D and Setting F}
    \label{fig:swd_df}
\end{figure}

Moreover, the average SWD for 
Setting F is 0.067, the same as 
Setting D. Further looking at 
the SWD values 
under different resolutions 
for the two different 
settings as shown in Fig. \ref{fig:swd_df}, 
it is clear that the two curves are very close. 
This validates our last inference, 
that when multiple layers are binarized 
together, the layer with least degree 
of redundancy is the bottleneck, which 
decides the overall performance of the network. 

\subsubsection{Experimental Results Using the Least Square Loss}
The experiments using the least square loss resulted 
are in accord with our previous experimental results as 
well as conclusions.

For each layer, the performance after binarization 
decreases along with the layer's redundancy.
The images generated by binarizing a single 
layer in the generator after training using the least 
square loss are shown in the Fig. \ref{fig:deconv_lsgan}. 
Same as in Fig. \ref{fig:deconv}, Fig. \ref{fig:orig_lsgan} 
displays the original results using the least square loss.
Figures \ref{fig:g1_lsgan} - \ref{fig:g4_lsgan} are 
generated by PBGen’s under Setting B - Setting E 
respectively using the least square loss. 
As mentioned before, the degree of redundancy 
of each layer is shown in Fig. \ref{fig:lvsredun}, 
which decreases along each layer. As a result, 
the quality of the generated images also decreases 
along binarizing each layer under Setting B, 
Setting C, Setting D, and Setting E. 
This is the same as that observed in 
the experiments based on 
the original loss function in DCGAN. 

The superposition of binarizability also holds 
in the experiments based on the least square loss.
The images generated by Setting 
F in Fig. \ref{fig:g1g2g3_lsgan}, 
where the first three layers in 
the generator are binarized together, 
show no significant difference 
from those in Figures \ref{fig:orig_lsgan}-
\ref{fig:g3_lsgan}. Binarizing any 
two layers from the first three 
layers (not shown here) 
will lead to a similar 
result. 
On the other hand, Setting 
G does not generate 
any meaningful output (Fig. 
\ref{fig:g1g2g3g4_lsgan}), as the last 
layer, which cannot be binarized 
alone, is binarized together 
with the first three layers. 
Binarizing any of the first 
three layers as well as the 
last layer (not shown here) 
will produce 
meaningless results too. 
Setting G follows 
the state-of-the-art binarization 
for CNNs directly without 
considering the degree 
of redundancy based on the least 
square loss. 
That is, with the assistance of 
the degree of redundancy, we can 
figure out that at most 
the first three 
deconvolution layers can 
be binarized with small loss 
on performance in the generator 
(Setting F) even based on the 
least square loss in DCGAN. 
Nevertheless, directly adopting 
the existing binarization method 
will lead to excessive degradation 
in performance and cannot provide 
any hint to improve (Setting G), 
even if a better loss function, 
the least square loss, is used.

\subsubsection{Compression Saving}
We also investigate the computation saving 
during training and inference 
and memory reduction of partially 
binarized deconvolution-based generators in 
hardware designs. Since BWN in \cite{rastegari2016xnor} 
is adopted to binarize layers, 
the same estimation on computation saving 
and memory cost as BWN is also utilized. 

Note that each binarized weight is 
$32\times$ small over its single 
precision presentation. Assume 
that out of a total of $N$ weights, 
$K$ are binarized. Then the new 
memory cost can be computed as 
\begin{equation}
\label{equ:memcost}
    (K+32\times(N-K))/(32\times N).
\end{equation}
On the other hand, \cite{rastegari2016xnor} 
mentioned the computation saving is 
$\sim2\times$ after binarization for 
a standard convolution operation, because 
multiplication is replaced by only 
addition and subtraction. 
This is also 
the situation when weights are binarized 
in a deconvolution operation, so the 
computation saving is adopted for a 
standard deconvolution operation.
That is, the new computation 
cost can also be calculated using 
(\ref{equ:memcost}) by replacing weights 
with deconvolution operations, and 
using $2$ instead of $32$.

\begin{table}[ht]
  \begin{center}
  \caption{Training and inference speedup as well as memory reduction for PBGen}
  \label{tlb:runtime}
  \begin{tabular}{c|c|c|c}
  \hline
  \multirow{2}{*}{\tabincell{c}{Generator model}} & \multicolumn{2}{|c|}{Computation Saving} & \multirow{2}{*}{\tabincell{c}{Memory\\Cost}} \\
  \cline{2-3}
  & Inference & Training & \\
  \hline
  \tabincell{c}{Original generator 
  from DCGAN\\(Setting A)} & 1.0$\times$ & 1.0$\times$ & 1.0$\times$ \\
  \hline
  \tabincell{c}{PBGen\\(Setting F)} & $\sim$1.96$\times$ & $\sim$1.32$\times$ & $\sim$1/25.81$\times$ \\
  \hline
  \end{tabular}
  \end{center}
\end{table}

TABLE \ref{tlb:runtime} summarizes the 
computation saving during training and 
inference as well as the memory 
reduction for PBGen compared with 
the original generator in DCGAN, 
which is the baseline model when 
considering the computation saving 
and the memory saving. 
PBGen under Setting F can achieve 
25.81$\times$ memory saving 
as well as 1.96$\times$ and 
1.32$\times$ speedup during 
inference and training 
respectively with 
little performance loss. 
For both the original 
generator and PBGen, 
during the training 
process the floating point representation 
of all weights need to be used for 
backward propagation and update 
\cite{rastegari2016xnor}. 
As such, the speedup mainly comes 
from faster forward propagation with 
binarized weights. 

\begin{figure}[h]
    \centering
    \includegraphics[width=3.2in]{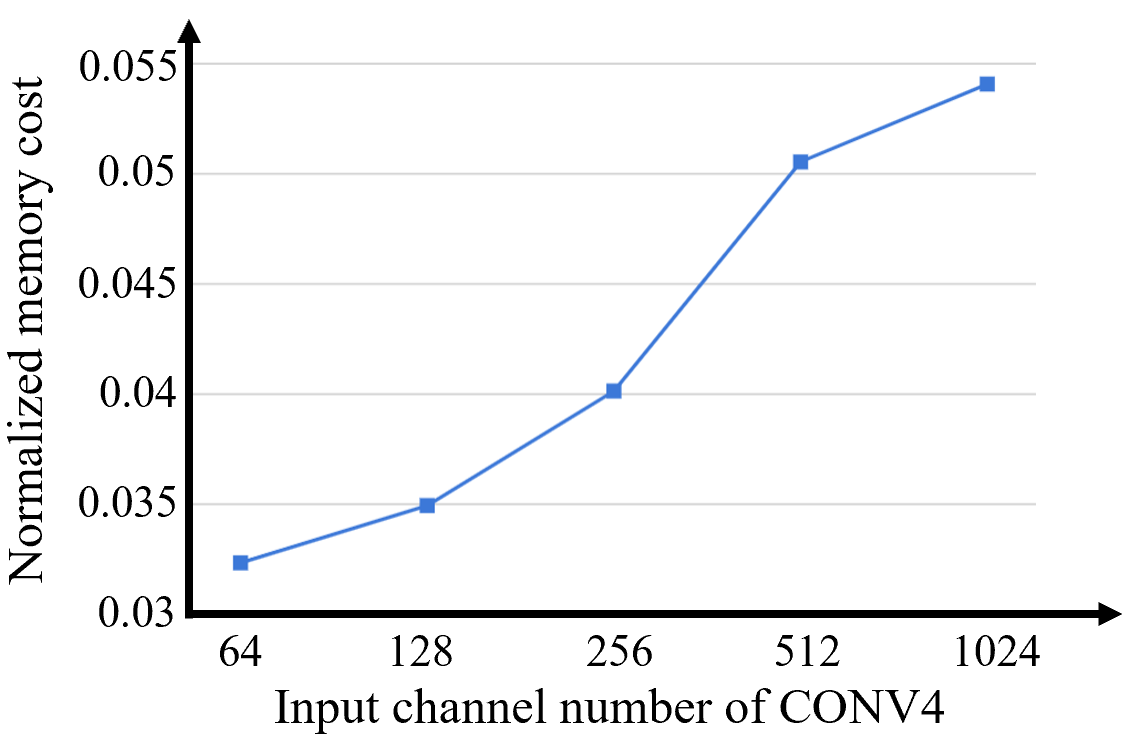}
    \caption{Memory cost (normalized to the original model without binarization) v.s. input channel number of CONV4.  Before the input channel number of CONV4 reaches 1024, only the first three 
deconvolution layers can be binarized without significant performance loss, and when the input channel number of CONV4 is 1024, all the four deconvolution 
layers can be binarized without significant loss.}
    \label{fig:mem2chn}
\end{figure}

The relationship between the memory 
saving and the input channel number of 
the fourth deconvolution layer (CONV4) 
on the generator is also investigated. 
Note that increasing the input channel 
number of the fourth convolution 
layer (CONV4) will increase its 
DOR and at the same time the memory 
cost.  On the other hand, eventually 
a high enough DOR (above 1024) will 
enable the layer to be binarized, 
leading to memory reduction. This can 
be seen in Fig. \ref{fig:mem2chn}, 
where x-axis is the input channel 
number of CONV4 and the y-axis is 
the total memory cost of the generator 
normalized to the original generator 
without any binarization. 
Before the input channel number 
of CONV4 reaches 1024, only the first three 
deconvolution layers can be binarized, so 
increasing DOR will result in the quick 
growth of memory cost. 
However, when the input channel 
number of CONV4 is 1024, all the four 
deconvolution layers can be binarized, 
which introduces extra memory saving to 
alleviate the memory cost increment.

\subsubsection{Unbalanced Competition} 
So far, our discussion has focused 
on the binarization of the 
generator in a 
GAN only, as the discriminator 
takes the same form as 
conventional CNNs. However, since 
competition between generator 
and discriminator 
is the key of GANs, would a binarized 
generator still compete well with a 
full discriminator?  

\begin{figure*}
    \centering
    \includegraphics[width=4.8in]{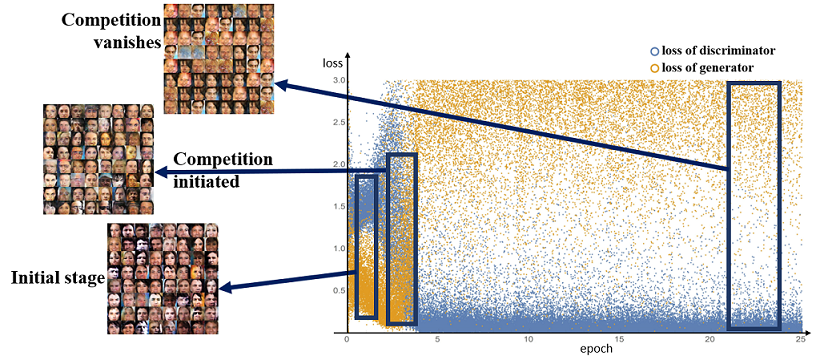}
    \caption{Loss values of the original 
    discriminator and PBGen under 
    Setting F along epochs}
    \label{fig:loss_g1g2g3}
\end{figure*}

\begin{figure*}
    \centering
    \includegraphics[width=4.8in]{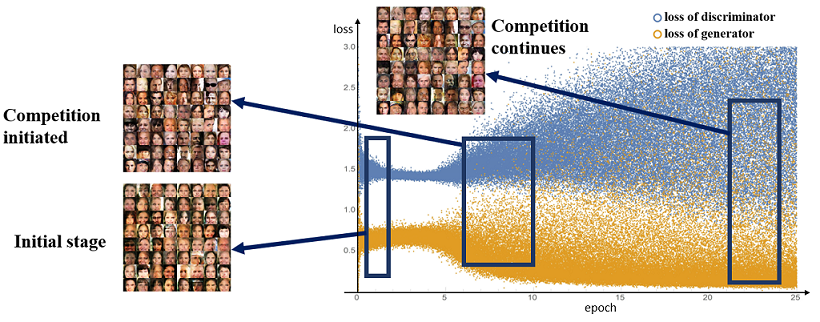}
    \caption{Loss values of binarized 
    discriminator and PBGen under Setting H 
    along epochs}
    \label{fig:loss_d4g3}
\end{figure*}

The loss values for the discriminator 
network and PBGen under 
Setting F are depicted 
in Fig. \ref{fig:loss_g1g2g3}, 
where x-axis indicates the number 
of epochs and y-axis is the loss value. 
The images generated from different 
number of epochs are also exhibited aside. 
From the figure we can see that 
during the initial stage, distorted 
faces are generated. As the competition 
is initiated, image quality improves. But 
very quickly, the competition vanishes, 
and the generated images stop improving. 
However, when we binarize the discriminator 
at the same time (Setting H), the 
competition continues to improve 
image quality, as can be 
seen in Fig. \ref{fig:loss_d4g3}. 

\begin{figure*}
    \centering
    \includegraphics[width=4.8in]{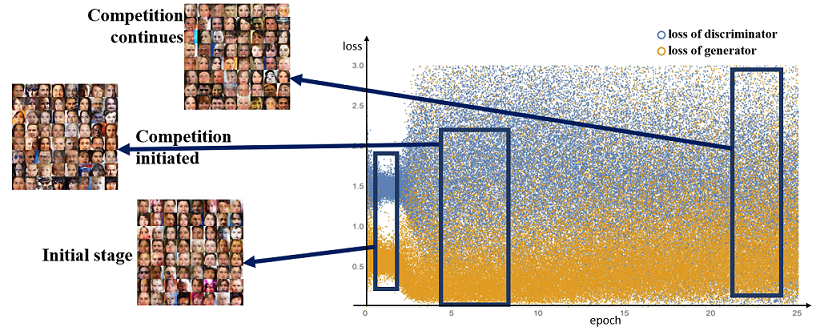}
    \caption{Loss values of the discriminator 
    and the generator in original DCGAN 
    under Setting A 
    along epochs}
    \label{fig:loss_orig}
\end{figure*}

We further plot the loss values of 
the discriminator and the generator 
of the original DCGAN (Setting A), 
and the results are shown in 
Fig. \ref{fig:loss_orig}. 
It is very similar to 
Fig. \ref{fig:loss_d4g3}, except 
that the competition is initiated 
earlier, which is due to the 
stronger representation power 
of both the generator and the 
discriminator before binarization. 
These figures confirm that the quick 
disappearance of competition is mainly 
due to the unbalanced generator 
and discriminator, which should be avoided. 

\section{Conclusion}
\label{sec:conc}
We now explore the quality of the images 
generated from balanced competition 
using Setting H. The images generated 
are shown in Fig. \ref{fig:d4g3} and 
Fig. \ref{fig:d4g3_lsgan}, 
the quality of which is apparently 
better than the rest 
in Fig. \ref{fig:deconv} and 
Fig. \ref{fig:deconv_lsgan} respectively. 
To further 
confirm this quantitatively, we compute 
the average SWD values of those images, 
which is 0.034 in 
average. This is even smaller than 
any average 
SWD values listed 
in TABLE \ref{tlb:swd}, which shows that 
the images are of better quality, even 
compared with the original DCGAN. 

\begin{figure}[htb]
    \centering
    \includegraphics[width=2.8in]{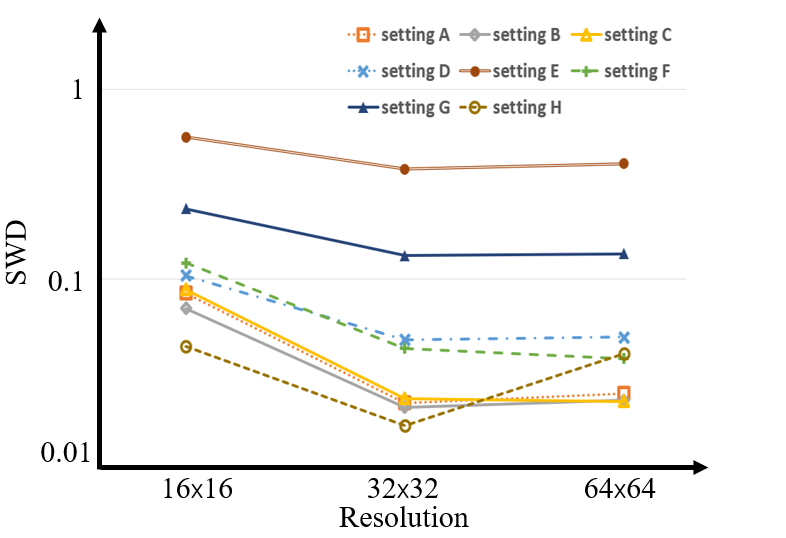}
    \caption{SWD v.s. resolutions 
    under all different settings}
    \label{fig:swd}
\end{figure}

\subsubsection{Summary}
To summarize the discussion and 
comparisons in this section, we 
plot the SWD v.s. resolution curves 
for all the 8 settings in 
Fig. \ref{fig:swd}. It allows a complete 
view of how these different 
settings compare in terms of 
similarity as a whole and fine details. 
From the figure we can see that 
Setting H gives the best similarity 
as a whole, 
while Setting C yields the 
finest detailed attributes.

Consequently, 
utilizing the degree of redundancy 
as a tool, 
we can efficiently find out eligible 
layers that can be binarized and 
based on their superposition, 
a final binarization 
strategy can be decided. 
It cannot guarantee an optimal 
result but does decrease the search 
space for the final solution 
from $O(2^n)$ to $O(n)$ or 
less, where 
$n$ is the number of layers, 
because testing on all combinations 
of binarization strategy is 
not necessary and 
we only need to binarize 
every single layer with 
high degree of redundancy to decide
the final strategy.
Since our theoretical analysis 
and experiments are based on 
deconvolutional layers, 
we believe this method can work 
for other deconvolution based 
generators beyond DCGAN. 

Compression techniques have been 
widely studied for convolutional 
neural networks, but directly adopting 
them to all layers will fail 
deconvolution-based generator 
in generative adversarial 
networks based on our observation. 
We propose and validate 
that the performance of deconvolution-based 
generator can be preserved when applying 
binarization to carefully selected layers 
(PBGen). To accelerate 
the process deciding whether a layer 
can be binarized or not, the degree of 
redundancy is proposed 
based on theoretical analysis 
and further verified by experiments. 
Under the guide of this metric, 
search space for optimal binarization 
strategy is decreased from $O(2^n)$ 
to $O(n)$ 
where $n$ is the number of 
layers in the generator. 
PBGen for DCGAN can yield up to 
25.81$\times$ saving in 
memory consumption with 1.96$\times$ 
and 1.32$\times$ speedup in inference 
and training respectively with 
little performance loss measured by 
sliced Wasserstein distance score. 
Besides, we also demonstrate that both 
generator and discriminator should 
be binarized at the same time 
for a balanced competition 
and better performance.

\ifCLASSOPTIONcaptionsoff
  \newpage
\fi


\bibliographystyle{IEEEtran}
\bibliography{jrnlbib}
%


%

\begin{IEEEbiography}{Michael Shell}
Biography text here.
\end{IEEEbiography}

\begin{IEEEbiographynophoto}{John Doe}
Biography text here.
\end{IEEEbiographynophoto}


\begin{IEEEbiographynophoto}{Jane Doe}
Biography text here.
\end{IEEEbiographynophoto}




\end{document}